\documentclass[runningheads]{llncs}
\usepackage{multirow}
\usepackage{makecell}
\usepackage{pifont}        
\usepackage{array}         
\usepackage{longtable}     
\usepackage{booktabs}      
\usepackage{calc}          
\usepackage{tabularx}      
\usepackage{float}         
\usepackage{amsmath,amssymb}
\usepackage{siunitx, tabularx, booktabs, makecell, multirow}
\usepackage[table]{xcolor}
\usepackage[T1]{fontenc}
\usepackage{graphicx}
\usepackage{blindtext}
\usepackage{hyperref}
\usepackage{marvosym}
\usepackage{microtype}
\usepackage{etoolbox}
\patchcmd{\thebibliography}{\clubpenalty4000}{\clubpenalty4000\setlength{\itemsep}{1pt}\setlength{\parsep}{1pt}}{}{}

\makeatletter
\renewcommand\section{\@startsection{section}{1}{\z@}%
                       {-2ex \@plus -0.5ex \@minus -.2ex}%
                       {0.8ex \@plus .2ex}%
                       {\normalfont\large\bfseries\rightskip\z@ \@plus 8em\pretolerance10000 }}
\renewcommand\subsection{\@startsection{subsection}{2}{\z@}%
                       {-1.5ex \@plus -0.5ex \@minus -.2ex}%
                       {0.5ex \@plus .2ex}%
                       {\normalfont\normalsize\bfseries\rightskip\z@ \@plus 8em\pretolerance10000 }}
\makeatother

\definecolor{bestblue}{HTML}{D9EFFF}
\colorlet{secondblue}{bestblue!30}
\definecolor{bestgreen}{HTML}{E6F7F4}
\colorlet{secondgreen}{bestgreen!30}
\definecolor{mix1}{HTML}{D7F3E9}

\usepackage[most]{tcolorbox}
\usepackage{xcolor}
\usepackage{pifont}
\usepackage{array}

\newtcolorbox{casebox}[2][]{
    enhanced,
    breakable,
    colback=white,
    colframe=blue!85!black,
    colbacktitle=blue!85!black,
    coltitle=white,
    title={#2},
    fonttitle=\bfseries,
    boxrule=0.8pt,
    arc=3mm,
    left=8pt,
    right=8pt,
    top=6pt,
    bottom=6pt,
    before skip=8pt,
    after skip=8pt,
    #1
}

\newcolumntype{L}[1]{>{\raggedright\arraybackslash}p{#1}}
\newcolumntype{R}[1]{>{\raggedleft\arraybackslash}p{#1}}
\newcolumntype{C}[1]{>{\centering\arraybackslash}p{#1}}

\newcommand{\orcidAuthorOne}	{\href{https://orcid.org/0009-0004-0679-2229}{\protect\includegraphics[scale=0.045]{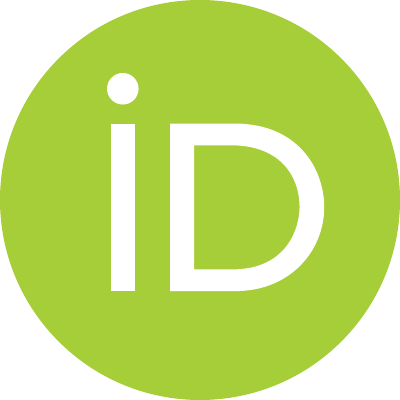}}}
\newcommand{\orcidAuthorTwo}	{\href{https://orcid.org/0000-0001-5499-4651}{\protect\includegraphics[scale=0.045]{img/orcid}}}
\newcommand{\orcidAuthorThree}	{\href{https://orcid.org/0000-0003-1179-4769}{\protect\includegraphics[scale=0.045]{img/orcid}}}
\newcommand{\orcidAuthorFour}	{\href{https://orcid.org/0009-0008-7266-6828}{\protect\includegraphics[scale=0.045]{img/orcid}}}
\newcommand{\orcidAuthorFive}	{\href{https://orcid.org/0000-0002-0929-9067}{\protect\includegraphics[scale=0.045]{img/orcid}}}
\newcommand{\orcidAuthorSix}	{\href{https://orcid.org/0000-0001-8673-9284}{\protect\includegraphics[scale=0.045]{img/orcid}}}

\titlerunning{Case-Based Adaptive LLM Tool Use}

\begin{document}

\title{Case-Based Calibration of Adaptive Reasoning and Execution for LLM Tool Use}
\author{Renning Pang\orcidAuthorOne \and
Tian Lan\orcidAuthorTwo \and 
Leyuan Liu(\Letter)\orcidAuthorThree \and 
Piao Tong\orcidAuthorFour \and
Sheng Cao\orcidAuthorFive \and
Xiaosong Zhang\orcidAuthorSix}
\authorrunning{Pang et al.}
\institute{University of Electronic Science and Technology of China, Chengdu 611731, China
\email{\{prn,piaot\}@std.uestc.edu.cn, \{lantian1029,leyuanliu,caosheng,johnsonzxs\}@uestc.edu.cn}}

\maketitle              
\begin{abstract}
Tool use extends large language models beyond parametric knowledge, but reliable execution requires balancing appropriate reasoning depth with strict structural validity. We approach this problem from a case-based perspective to present CAST, a case-driven framework that treats historical execution trajectories as structured cases. Instead of reusing raw exemplar outputs, CAST extracts case-derived signals to identify complexity profiles for estimating optimal reasoning strategies, alongside failure profiles to map likely structural breakdowns. The framework translates this knowledge into a fine-grained reward design and adaptive reasoning, enabling the model to autonomously internalize case-based strategies during reinforcement learning. Experiments on BFCLv2 and ToolBench demonstrate that CAST improves both schema-faithful execution and task-level tool-use success while reducing unnecessary deliberation. The approach achieves up to 5.85 percentage points gain in overall execution accuracy and reduces average reasoning length by 26\%, significantly mitigating high-impact structural errors. Ultimately, this demonstrates how historical execution cases can provide reusable adaptation knowledge for calibrated tool use.

\end{abstract}

\section{Introduction}\label{sec:intro}
Tool use has become a central mechanism for extending large language models (LLMs) beyond parametric knowledge, and recent surveys identify tool invocation, planning, and action coordination as core components of LLM-agent systems \cite{ruan2023tptu,chen2025toollearning,chowa2025language2action,yehudai2025agenteval}. At the same time, benchmark-oriented studies suggest that reliable function calling remains far from solved, especially in settings that require memory, dynamic decision-making, and long-horizon reasoning rather than single-turn text generation alone \cite{patil2025bfcl}. Representative frameworks such as Toolformer \cite{schick2023toolformer}, ReAct \cite{yao2023react}, and ToolLLM \cite{qin2023toollm} further show that the key challenge is no longer simply whether LLMs can call tools, but whether they can allocate reasoning effort and execute structured tool actions in a stable and task-sensitive manner.

This challenge arises because tool use places two simultaneous demands on the model. Different queries require different amounts of intermediate reasoning: some can be solved with little deliberation, whereas others require additional reasoning to verify constraints, normalize arguments, or compose multi-step tool calls. Recent work on adaptive chain-of-thought (CoT) and efficient tool calling suggests that treating all inputs with the same reasoning policy is often inefficient and can even be counterproductive \cite{ma2025cotvalve,lou2025adacot,xu2025alignment}. At the same time, tool execution is governed by strict structural constraints, while standard reinforcement learning (RL) often provides only coarse end-task feedback, making it difficult to identify whether failure originates from tool selection, parameter coverage, type mismatch, schema violation, or value construction. The central difficulty in tool use is therefore to calibrate reasoning depth and execution structure jointly under heterogeneous task demands.

\begin{figure}[t]
\centering
\includegraphics[width=0.7\columnwidth]{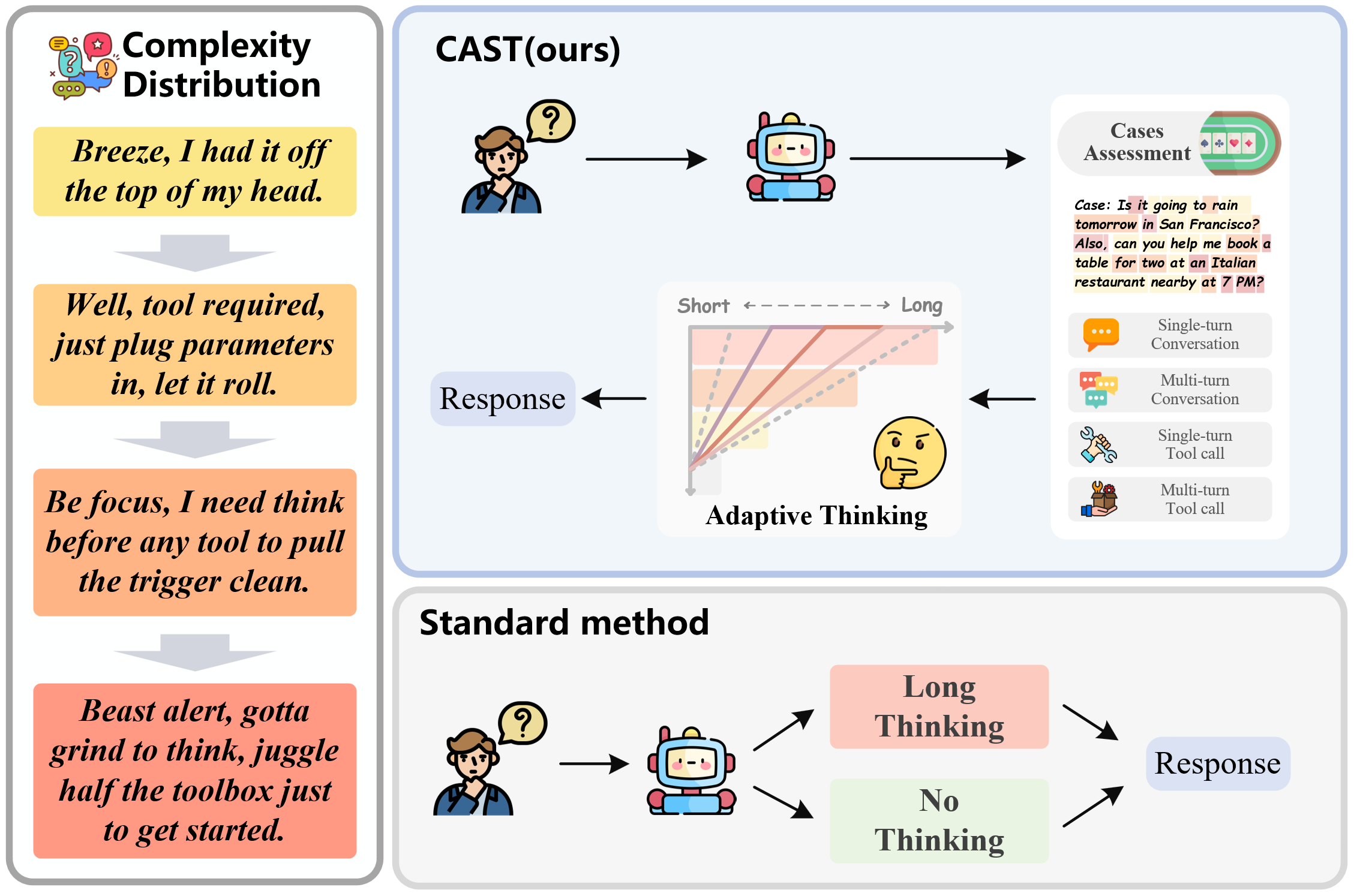} 
\caption{Comparison of reasoning workflows with and without a case-based mechanism.}
\label{fig1}
\end{figure}

We approach this problem from a modern Case-Based Reasoning (CBR) perspective on tool-use adaptation in LLMs. In our setting, historical execution trajectories are treated as cases that record how a task was solved, how much reasoning preceded action, whether the tool invocation succeeded, and what kinds of failures arose when it did not. This view is consistent with classical CBR \cite{aamodt1994cbr} and with subsequent work emphasizing adaptation bottlenecks and case-base maintenance \cite{hanney1997adaptation,craw2006adaptation}. Recent studies further suggest that LLMs can support several stages of the CBR process, including case adaptation, similarity assessment, and experience-grounded agent reasoning \cite{lenz2023argument,lenz2025llsim,liu2025offline2online,bergmann2025exar}. Together, these developments motivate the use of historical execution cases as structured sources of calibration knowledge for tool use. In this paper, we formulate tool use as a case-based adaptation problem and introduce \textbf{C}ase-driven \textbf{A}daptation for \textbf{S}chema-faithful \textbf{T}ool use (CAST), a case-driven framework that extracts two forms of case-derived signals from past trajectories. Specifically, it derives a complexity profile to estimate the necessary reasoning depth and a failure profile to map likely structural breakdowns. As shown in Figure\ref{fig1}, rather than imposing a static reasoning limit or a uniform chain-of-thought policy, CAST translates this case-derived knowledge into a fine-grained reward design that supports adaptive reasoning. In particular, the model learns to shorten deliberation for easy cases while preserving sufficient reasoning steps for cases that require constraint verification, argument normalization, or multi-step tool composition. This approach enables the model to internalize historical case experiences during reinforcement learning, empowering it to autonomously orchestrate the model's reasoning budget and execute schema-faithful tool actions. Experiments on BFCLv2 and ToolBench show that CAST improves both schema-faithful execution and task-level tool-use success while reducing unnecessary reasoning, with especially strong gains on more complex cases and clear reductions in high-impact structural failures.

The remainder of this paper is organized as follows. Section~\ref{sec:related} reviews related work on CBR for LLMs and agents, adaptive reasoning, and tool-use alignment. Section~\ref{sec:problem} formulates tool use as a case-based adaptation problem. Section~\ref{sec:method} presents the CAST framework. Section~\ref{sec:exp} reports the experimental results and analyses. Section~\ref{sec:conclusion} concludes the paper and discusses limitations and future work.

\section{Related Work}\label{sec:related}
Case-based reasoning (CBR) addresses new problems by reusing and adapting solutions from similar prior cases through retrieval, reuse, revision, and retention \cite{aamodt1994cbr}. In recent years, this perspective has increasingly intersected with large language models and agentic systems. Prior work has explored the use of LLMs for case adaptation in argumentative reasoning \cite{lenz2023argument}, while more recent studies investigate how LLMs can support similarity assessment in case retrieval \cite{lenz2025llsim}. Other work extends case-based ideas to reinforcement learning and agent settings, for example through case-based knowledge distillation for reinforcement learning \cite{liu2025offline2online}, and unified experience-grounded agentic reasoning architectures \cite{bergmann2025exar}. Recent work has also examined how LLMs can support case-base population from unstructured sources \cite{ghazouani2025casebase} and how CBR can be integrated with LLMs in practical decision-support settings such as fraud detection \cite{ge2025expensefraud}. Taken together, these studies suggest that LLMs can support multiple stages of the CBR process, including similarity assessment, adaptation, case acquisition, and the use of structured prior experience in agent reasoning.

A related line of research studies how to regulate the amount of reasoning generated by large language models. Existing methods seek to shorten, compress, or selectively trigger chain-of-thought in order to improve efficiency while maintaining competitive task performance \cite{ma2025cotvalve,lou2025adacot}. This line of work is directly relevant to our setting because it recognizes that different inputs may require different amounts of intermediate reasoning. In parallel, another line of work focuses on tool use and reasoning--acting behavior in LLMs. Toolformer shows that language models can learn when and how to invoke external tools \cite{schick2023toolformer}, while ReAct demonstrates the value of interleaving reasoning traces with external actions \cite{yao2023react}. Tool-use research has also been accompanied by increasingly realistic evaluation settings, including large-scale API benchmarks and function-calling leaderboards \cite{qin2023toollm,patil2025bfcl}. More recent work further improves invocation discipline and efficiency, for example through meta-cognitive triggering \cite{li2025adaptive}, alignment for efficient tool calling \cite{xu2025alignment}, or token-level policy gradient reshaping for tool-use LLMs \cite{Lin2026ResT}. These studies substantially advance reasoning control and tool-use alignment, but they generally treat reasoning depth and execution structure as separate concerns rather than as jointly case-conditioned aspects of the same problem.

Our work is motivated by the gap between these directions. Existing CBR-oriented research suggests that historical cases can provide reusable structure for future reasoning and action, while adaptive reasoning and tool-alignment research shows that both reasoning cost and execution discipline must be carefully controlled. What remains underexplored is how historical execution cases can be used to derive calibration signals for these two aspects jointly. CAST is a case-driven framework in which historical execution cases provide case-derived complexity and failure signals, designed to explicitly internalize the case-based experience that calibrates reinforcement learning through reasoning-budget control and schema-faithful tool optimization.

\section{Problem Formulation}\label{sec:problem}
Given a user query $q$ and a tool set $\mathcal{T}$, tool use requires a model to generate a trajectory $\tau=(r,c)$, where $r$ denotes the reasoning trace and $c=\mathrm{call}(f,z)$ denotes a structured tool invocation with function $f$ and arguments $z$. We formulate this problem from a case-based adaptation perspective: historical execution trajectories are treated as execution cases, $\xi_i=(q_i,r_i,c_i,o_i,\phi_i)$, where $q_i$ is the input query, $r_i$ is the reasoning trace, $c_i$ is the executed tool call, $o_i$ is the execution outcome, and $\phi_i=(h_i,e_i)$ is the case profile consisting of a complexity profile $h_i$ and a failure profile $e_i$. Under this formulation, historical executions are not treated merely as supervision traces, but as structured cases from which case-derived calibration signals can be obtained for reasoning-budget control and schema-faithful tool use. The objective is therefore to learn a policy $\pi_\theta$ calibrated by these historical cases, such that the generated trajectory $\tau \sim \pi_\theta(\cdot \mid q,\mathcal{T})$ produces tool actions that are both semantically appropriate and structurally executable.

\begin{figure}[t]
\centering
\includegraphics[width=\textwidth]{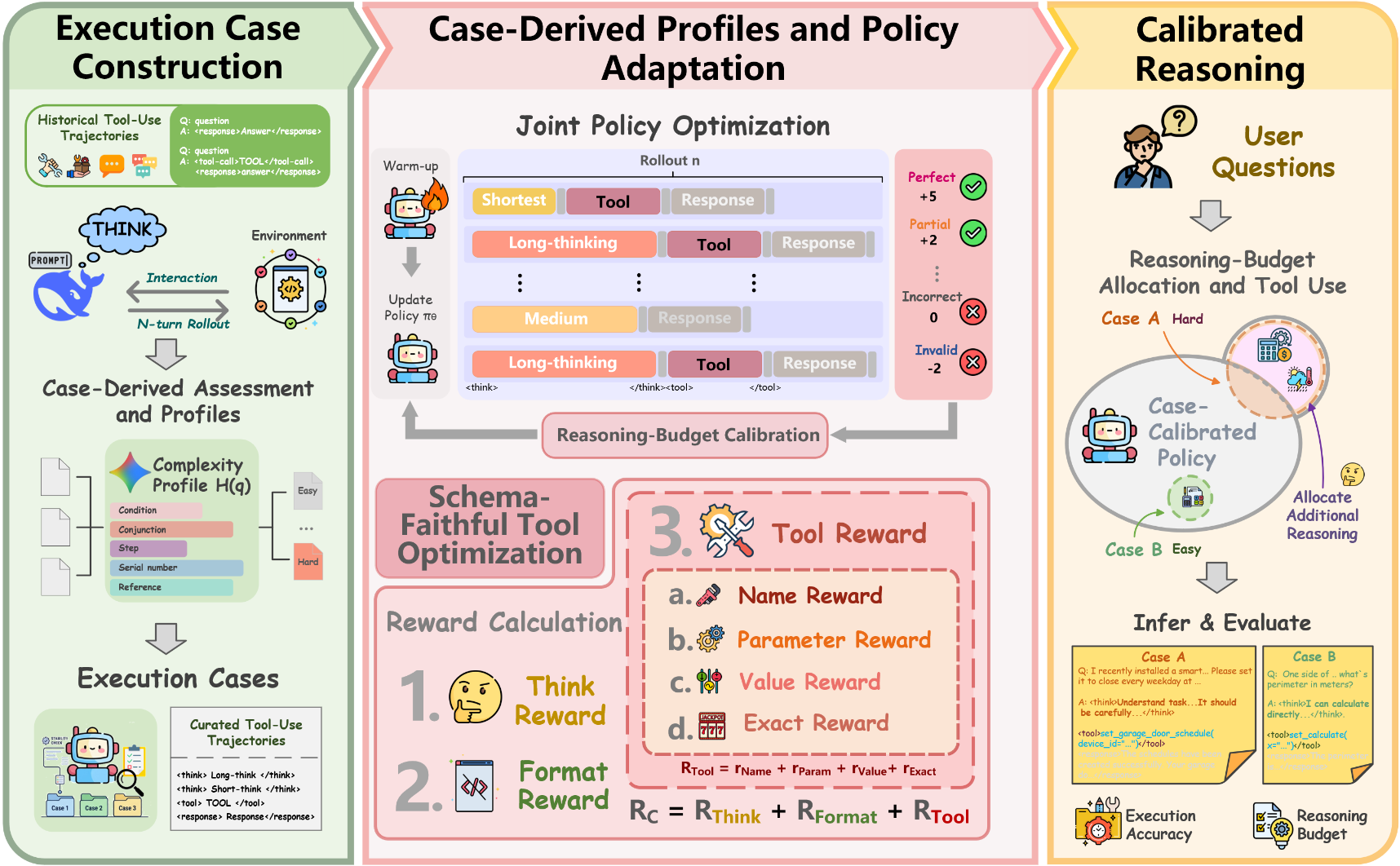}
\caption{Overview of CAST. Historical tool-use trajectories are organized as execution cases. These case signals guide two coordinated adaptation processes: reasoning-budget calibration and schema-faithful tool optimization. Reinforcement learning serves as the optimization mechanism for this case-based adaptation framework.}
\label{fig2}
\end{figure}

\section{Method}\label{sec:method}
Figure~\ref{fig2} illustrates the overall architecture of CAST. From a case-based perspective, CAST organizes historical tool-use trajectories as structured execution cases, derives case-derived complexity and failure signals from these cases, and uses them to guide two coordinated adaptation processes: reasoning-budget calibration and schema-faithful tool optimization. In implementation, CAST is trained with a supervised warm-up stage followed by reinforcement learning calibrated by these case-derived signals. In this sense, reasoning-budget calibration model is encouraged to vary its deliberation depth according to case complexity before producing a schema-faithful tool action. Under this formulation, reinforcement learning serves as the optimization mechanism, while the methodological core lies in case-driven capability assessment and adaptation.

\subsection{Execution Case Construction}\label{sec:case_const}
CAST first converts historical tool-use trajectories into execution cases. Unlike ordinary supervision pairs, an execution case preserves not only the input query and target tool call, but also the intermediate reasoning process and the observed execution outcome. This allows historical trajectories to function as structured experience rather than isolated input--output examples. Formally, each case is represented as $\xi_i = (q_i, r_i, c_i, o_i, \phi_i)$, where $q_i$ is the user query, $r_i$ is the reasoning trace, $c_i$ is the structured tool invocation, $o_i$ is the execution outcome, and $\phi_i$ is the case profile defined in Section~\ref{sec:case_rep}. In practice, we bootstrap the case base from tool-augmented reasoning trajectories distilled from DeepSeek-R1, normalize all tool interactions into a unified function-schema format, and filter malformed or unverifiable traces during preprocessing.

\subsection{Case Representation: Complexity and Failure Profiles}\label{sec:case_rep}
After constructing the execution case base, CAST represents each case through a profile $\phi_i=(h_i,e_i)$, where $h_i$ is a complexity profile and $e_i$ is a failure profile. The complexity profile characterizes how much intermediate reasoning is typically required before action, whereas the failure profile characterizes where execution is most likely to break, such as function selection, argument construction, type mismatch, or schema violation. Together, these profiles provide the case-derived calibration signals used in the subsequent adaptation stages.

We estimate case complexity from the base model's observed execution behavior rather than from a manually specified heuristic. For each historical query-target pair $(q,a)$, we sample a trajectory $y \sim \pi_{\text{base}}(\cdot \mid q)$ and evaluate it with the verifier $V$. If the generated tool call is fully correct and schema-faithful, the case is assigned zero hardness. Otherwise, the failed trajectory is assessed by an external rubric-guided judge using reference examples of simple and difficult cases. In our implementation, this judge is instantiated with Gemini-Pro. We define the resulting hardness score as
\begin{equation}
H(q)=
\begin{cases}
0, & V(y,a)=1,\\[4pt]
1-S_{\mathrm{judge}}(q,y), & V(y,a)=0,
\end{cases}
\end{equation}
where $S_{\mathrm{judge}}(q,y)\in[0,1]$ denotes the judged adequacy of the failed trajectory with respect to logical coherence, parameter construction, and schema adherence. We use this score as an operational proxy for reasoning demand, and partition the case base into $D_{\mathrm{easy}}$ and $D_{\mathrm{hard}}$ for curriculum scheduling while retaining the continuous score for finer-grained analysis.

The failure profile is derived from the mismatch between the generated tool call and the reference execution. For each failed case, we record the principal failure dimensions involved in the mismatch, including function-name errors, parameter-key omissions, type mismatches, constraint violations, and value mismatches, and summarize them as
\begin{equation}
e_i=(e_i^{\mathrm{name}},e_i^{\mathrm{key}},e_i^{\mathrm{type}},e_i^{\mathrm{constraint}},e_i^{\mathrm{value}}).
\end{equation}
Each component indicates whether the corresponding failure pattern is present in the trajectory.

\subsection{Case-Based Adaptation for Reasoning-Budget Calibration}\label{sec:reasoning_calib}
The adaptation problem in tool use is to determine how much reasoning should precede action for a given case. A uniform reasoning policy is inadequate because different cases require different amounts of intermediate deliberation: simple cases may suffer from unnecessary verbosity. CAST addresses this problem by using the case complexity profile to internalize historical execution strategies into the model's policy. Through a fine-grained reward design, the model learns to autonomously produce appropriately short or long CoT traces, mirroring the successful adaptation patterns found in similar prior cases.

Concretely, the complexity profile introduced in Section~\ref{sec:case_rep} is operationalized by the hardness score $H(q)\in[0,1]$, where smaller values indicate easier cases and larger values indicate more difficult ones. We discretize $H(q)$ into a small number of difficulty bands $d(q)$ only for estimating stable empirical length baselines, while retaining the continuous value $H(q)$ itself as the control signal for reward shaping. Let $r_a \in [-2,2]$ denote the final answer score and $L$ the response length. We define a difficulty-conditioned length baseline as:
\begin{equation}
L_{emp}^{d(q)}(t) = L_{max}^{d(q)} - \bigl(L_{max}^{d(q)} - L_{target}^{d(q)}\bigr)\cdot \min\left(1, \frac{t}{T_{warmup}}\right),
\end{equation}
where $L_{max}^{d(q)}$ is the initial relaxed length limit for difficulty band $d(q)$, $L_{target}^{d(q)}$ is the target concise length, and $T_{warmup}$ is the curriculum duration.

We then define the excess-length ratio
\begin{equation}
\rho(q,L)=\max\!\left(0,\frac{L}{L_{emp}^{d(q)}}-1\right),
\end{equation}
which measures how much the current response exceeds the expected reasoning budget, and the complexity-sensitive gating weight
\begin{equation}
\lambda(q)=1-H(q),
\end{equation}
so that easy cases are more strongly penalized for unnecessary overthinking, whereas difficult cases are less sensitive to length and remain primarily correctness-driven.

The shaping coefficient is defined as
\begin{equation}
\alpha(q,r_a,L)=
\begin{cases}
\max\!\left(0,\,1-\lambda(q)\rho(q,L)\right), & r_a>0,\\[6pt]
1+\lambda(q)\rho(q,L), & r_a<0,\\[6pt]
1, & r_a=0,
\end{cases}
\end{equation}
and the reasoning-side reward is
\begin{equation}
\mathcal{R}_{\mathrm{Think}} = \alpha(q,r_a,L)\cdot r_a.
\end{equation}

Under this formulation, the same complexity signal plays two roles. Through $d(q)$, it provides a stable empirical baseline for expected reasoning length; through $H(q)$, it determines how strongly overlong reasoning should be penalized. As a result, easy cases are encouraged to remain concise, while difficult cases preserve a larger reasoning workspace and are optimized primarily with respect to answer correctness.

\subsection{Failure-Profile-Grounded Optimization for Schema-Faithful Tool Use}\label{sec:tool_opt}
While reasoning-budget calibration governs how much deliberation precedes action, a complementary challenge remains: ensuring that the final tool action is structurally executable. Even when a reasoning trace appears semantically plausible, execution can still fail because of errors in function names, argument keys, types, constraints, or values. CAST therefore treats tool optimization as a separate but coordinated objective grounded in the failure profile $e_i=\bigl(e_i^{\mathrm{name}}, e_i^{\mathrm{key}}, e_i^{\mathrm{type}}, e_i^{\mathrm{constraint}}, e_i^{\mathrm{value}}\bigr)$. Let $\mathcal{G}$ and $\mathcal{P}$ denote the ground-truth and predicted collections of tool calls. To handle multi-call settings, we align them by maximum-weight bipartite matching:

\begin{equation}
    \mathcal{J}=\arg\max_{M\in\mathcal{M}(\mathcal{G},\mathcal{P})}\sum_{(G,P)\in M}s_{\mathrm{match}}(G,P),
\end{equation}

where $s_{\mathrm{match}}(G,P)=\delta(\mathrm{name}(G),\mathrm{name}(P))+\frac{|K^G\cap K^P|}{|K^G\cup K^P|}$, and $K^G,K^P$ are the parameter-key sets of $G,P$. Repeated tool calls remain distinct nodes in the matching graph.

We then define a six-dimensional structural reward vector:
\begin{equation}
\mathbf{r}_{\mathrm{tool}}=(r_{\mathrm{name}}, r_{\mathrm{key}}, r_{\mathrm{type}}, r_{\mathrm{constraint}}, r_{\mathrm{value}}, r_{\mathrm{exact}})^\top.
\end{equation}
Let $N_G,N_P$ be the overall function-name sets of $\mathcal{G},\mathcal{P}$. The name score is the Jaccard overlap $r_{\mathrm{name}}=\frac{|N_G\cap N_P|}{|N_G\cup N_P|}$. The key score averages parameter-key overlap over aligned calls: $r_{\mathrm{key}}=\frac{1}{|\mathcal{J}|}\sum_{j\in\mathcal{J}}\frac{|K_j^G\cap K_j^P|}{|K_j^G\cup K_j^P|}$ when $|\mathcal{J}|>0$, and $0$ otherwise; when $K_j^G=K_j^P=\varnothing$, the corresponding overlap term is set to $1$. We define $r_{\mathrm{type}}, r_{\mathrm{constraint}}, r_{\mathrm{value}}$ as the average indicator matches over overlapping keys, with default value $0$ when the number of overlapping keys is zero. The exact-match term is $r_{\mathrm{exact}}=V_{\mathrm{AST}}(\mathcal{P},\mathcal{G})\in\{0,1\}$, where AST denotes Abstract Syntax Tree matching. The raw structural score is:
\begin{equation}
R_{\mathrm{raw}}=
r_{\mathrm{name}}+
r_{\mathrm{key}}+
r_{\mathrm{type}}+
r_{\mathrm{constraint}}+
r_{\mathrm{value}}+
r_{\mathrm{exact}}.
\end{equation}

Since each component is bounded in $[0,1]$, the global maximum is $S_{\max}=6$. We therefore define the final tool-side reward as
\begin{equation}
\mathcal{R}_{\mathrm{Tool}}=2\cdot \frac{R_{\mathrm{raw}}}{S_{\max}}-1 \in [-1,1].
\end{equation}

As a result, CAST provides dense and interpretable credit assignment for schema-faithful tool use without hand-tuned gating rules or query-dependent scaling constants.

\subsection{Training Objective and Optimization}
Given a query $q$ and a tool set $\mathcal{T}$, the model generates a trajectory $\tau=(r,c)$. To guide the optimization securely and prevent degenerate behaviors, we formulate a composite reward that provides feedback at three complementary levels of abstraction:
\begin{equation}
\mathcal{R}_{C} = \mathcal{R}_{\mathrm{Think}} + \mathcal{R}_{\mathrm{Format}} + \mathcal{R}_{\mathrm{Tool}}.
\end{equation}
The first term, $\mathcal{R}_{\mathrm{Think}}$, is derived from the complexity profile and aligns the model toward adaptive reasoning by calibrating the intermediate deliberation budget for each case. (Section~\ref{sec:reasoning_calib}). The second, $\mathcal{R}_{\mathrm{Format}}$, is a rule-based guardrail that enforces valid tag encapsulation. The third, $\mathcal{R}_{\mathrm{Tool}}$, relies on the failure profile to evaluate fine-grained schema-faithful accuracy (Section~\ref{sec:tool_opt}).

In practice, we first initialize the model using SFT on curated tool-augmented trajectories with the standard next-token prediction loss:
\begin{equation}
    \mathcal{L}_{\mathrm{SFT}}
= - \frac{1}{|\mathcal{D}_{\mathrm{SFT}}|}
\sum_{(x,y)\in \mathcal{D}_{\mathrm{SFT}}}
\sum_{t=1}^{|y|}
\log \pi_{\theta}(y_t \mid x, y_{<t}).
\end{equation}
After initialization, we optimize the policy $\pi_{\theta}$ using a Group Relative Policy Optimization (GRPO)-based procedure augmented with the case-derived composite reward $\mathcal{R}_C$, which replaces the standard scalar reward with our decomposed complexity- and failure-conditioned signals. To stabilize the learning dynamics, we organize the RL training following an Easy-to-Hard curriculum based on the difficulty subsets partitioned by the complexity profile. For each query $q$, we sample a group of $G$ trajectories $\{\tau_1, \dots, \tau_G\}$. The group-relative advantage for the $i$-th trajectory is computed by normalizing the composite rewards within the group:
\begin{equation}
A_i = \frac{\mathcal{R}_C(\tau_i) - \mu_{\mathcal{R}}}{\sigma_{\mathcal{R}} + \epsilon_{\text{stab}}},
\end{equation}
where $\mu_{\mathcal{R}}$ and $\sigma_{\mathcal{R}}$ are the mean and standard deviation of the group rewards, and $\epsilon_{\text{stab}}$ is a small constant for numerical stability. The policy is then updated by maximizing the clipped surrogate objective augmented with a KL divergence penalty:
\begin{multline}
\mathcal{J}_{\mathrm{CAST}}(\theta) = \mathbb{E}_{q, \{\tau_i\}_{i=1}^G} \bigg[ \min\Big( \rho_i A_i, \mathrm{clip}\big( \rho_i, 1-\epsilon, 1+\epsilon \big) A_i \Big) \\
- \beta \cdot D_{\mathrm{KL}}\Big( \pi_\theta(\cdot \mid q) \parallel \pi_{\mathrm{ref}}(\cdot \mid q) \Big) \bigg],
\end{multline}
where $\rho_i = \frac{\pi_\theta(\tau_i \mid q)}{\pi_{\theta_{\text{old}}}(\tau_i \mid q)}$ is the probability ratio, $\epsilon$ is the clipping margin, $\beta$ controls the KL penalty strength, and $\pi_{\mathrm{ref}}$ is the reference policy. Under this formulation, SFT provides warm-up initialization and the GRPO-based procedure serves as the optimization engine, while the methodological core of CAST lies in representing historical executions as structured cases, deriving complexity and failure profiles, and translating them into the decomposed case-derived signals that drive the GRPO updates.

\section{Experiments}\label{sec:exp}
\subsection{Experimental Setup}\label{sec:setup}
We evaluate CAST on BFCLv2, including 5,551 instances covering single-turn, parallel, multi-step, and irrelevance-detection scenarios, and on ToolBench with diverse queries mapping to REST APIs across 49 categories. Model-level baselines include the original backbones alongside their SFT and GRPO variants. We primarily use Qwen2.5-7B-Instruct, adding Qwen2.5-Coder-7B-Instruct and Llama-3.2-8B-Instruct to examine cross-backbone robustness, and report closed models (GPT-4o \cite{Hurst2024GPT4oSC}, Qwen-Max \cite{Yang2024Qwen25TR}, DeepSeek-V3 \cite{DeepSeek-AI2024DeepSeekV3TR}) for context. Method-level comparisons feature Toolformer~\cite{schick2023toolformer}, ReAct~\cite{yao2023react}, ToolAlign~\cite{chen2024toolalign}, CoT-Valve~\cite{ma2025cotvalve}, OTC~\cite{wang2025otc}, Granite~\cite{abdelaziz2024granite}, and Gorilla~\cite{patil2023gorilla} evaluated under identical protocols. We utilize CAST via Megatron with an 8K maximum response length and evaluate inference using SGLang. To balance exploration and warm-up, we train for 2 epochs during the SFT phase. For RL training, we sample a group of $G=8$ rollouts per query at a temperature of $0.9$. The policy is optimized using AdamW with a peak learning rate of $1 \times 10^{-6}$, a cosine learning rate scheduler, and a weight decay of $0.01$. To stabilize the reinforcement learning dynamics, we apply a KL divergence penalty coefficient of $\beta=0.01$.

\begin{table}[ht]
  \centering
  \scriptsize
  \caption{Main results on BFCLv2 and ToolBench. We report execution accuracy on BFCLv2 (Non-Live AST, Live, and Overall) and task success on ToolBench (Pass and Win).}
  \label{tab:bfcl_toolbench_all}
  \begin{tabular}{llccccc}
    \toprule
    \textbf{Category} & \textbf{Model / Method} & \textbf{Non-Live AST} & \textbf{Live} & \textbf{Overall} & \textbf{Pass} & \textbf{Win} \\
    \cmidrule(lr){3-5}\cmidrule(lr){6-7}
    & & \multicolumn{3}{c}{\textbf{BFCLv2 (\%)}} & \multicolumn{2}{c}{\textbf{ToolBench (\%)}} \\
    \midrule
    \multirow{3}{*}{\makecell[l]{\textit{Closed} \\ \textit{Source}}}
      & GPT-4o                   & 86.83 & 78.92 & 82.88 & 64.43 & 67.63 \\
      & Qwen-Max                 & 84.97 & 80.85 & 82.91 & 70.93 & 71.83 \\
      & DeepSeek-V3              & 86.37 & 75.28 & 80.83 & 70.73 & 72.93 \\
    \midrule
    \multirow{9}{*}{\makecell[l]{\textit{Open} \\ \textit{Source}}}
      & Llama-3.2-8B-Instruct-SFT   & 83.14 & 73.82 & 78.48 & 70.63 & 45.83 \\
      & Llama-3.2-8B-Instruct-GRPO  & 82.18 & 75.21 & 78.73 & 71.23 & 45.23 \\
      & Llama-3.2-8B-Instruct-CAST  & 83.95 & 76.84 & 80.43 & 75.93 & 47.23 \\
      \cmidrule(lr){2-7}
      & Qwen2.5-Coder-7B-Instruct-SFT   & 86.07 & 74.92 & 80.86 & 67.84 & 65.37 \\
      & Qwen2.5-Coder-7B-Instruct-GRPO  & 86.51 & 75.07 & 80.79 & 71.13 & 67.82 \\
      & Qwen2.5-Coder-7B-Instruct-CAST  & 87.12 & 82.43 & 84.70 & 80.61 & 79.14 \\
      \cmidrule(lr){2-7}
      & Qwen2.5-7B-Instruct-SFT    & 86.23 & 78.90 & 82.58 & 68.67 & 65.23 \\
      & Qwen2.5-7B-Instruct-GRPO   & 87.05 & 80.29 & 83.67 & 72.71 & 68.23 \\
      & Qwen2.5-7B-Instruct-CAST   & 88.24 & 87.40 & 88.43 & 80.67 & 79.43 \\
    \midrule
    \multirow{6}{*}{\textit{Other}}
      & Granite    & 86.17 & 79.19 & 84.71 & 68.47 & 50.17 \\
      & Gorilla    & 86.02 & 80.44 & 82.21 & 62.27 & 46.29 \\
      & Toolformer & 76.11 & 59.47 & 67.07 & 48.92 & 22.11 \\
      & ReAct      & 73.58 & 58.43 & 66.08 & 43.37 & 18.22 \\
      & ToolAlign  & 77.26 & 61.47 & 71.16 & 46.78 & 22.36 \\
      & OTC        & 82.64 & 71.28 & 78.33 & 65.49 & 36.12 \\
    \bottomrule
  \end{tabular}
\end{table}

\subsection{Overall Performance and Efficiency}\label{sec:results}
Table~\ref{tab:bfcl_toolbench_all} reports the main results on BFCLv2 and ToolBench. On BFCLv2, CAST consistently improves over both SFT and GRPO across all three open-source backbones, indicating that case-derived calibration transfers across different model families and pretraining styles. The strongest result is obtained with Qwen2.5-7B-Instruct, where CAST reaches 88.43\% overall, gaining 5.85 points over SFT and 4.76 points over GRPO. The narrow 0.84-point gap between Non-Live AST and Live scores suggests that the gains transfer from offline schema matching to actual tool execution. The same trend is visible on the other open-source backbones. On Llama-3.2-8B-Instruct, CAST improves BFCLv2 Overall from 78.48\% under SFT and 78.73\% under GRPO to 80.43\%. On Qwen2.5-Coder-7B-Instruct, CAST reaches 84.70\% overall, with especially large gains on Live execution, rising from 74.92\% and 75.07\% to 82.43\%. Together, these results suggest that CAST is particularly effective when evaluation places stricter weight on execution validity. On ToolBench, the transfer pattern is also consistently positive. CAST achieves the strongest ToolBench results among CAST-trained models with Qwen2.5-7B-Instruct, reaching 80.67\% Pass and 79.43\% Win, both substantially above its SFT and GRPO counterparts. Llama-3.2-8B-Instruct shows a similar trend. Notably, the updated results on Qwen2.5-Coder-7B-Instruct now reveal a strong monotonic improvement as well: ToolBench Pass increases from 67.84\% under SFT and 71.13\% under GRPO to 80.61\% under CAST, while Win rises from 65.37\% and 67.82\% to 79.14\%. Rather than indicating a trade-off between schema-faithful calibration and end-to-end task success, the coder backbone now provides additional evidence that case-derived calibration transfers beyond local structural correctness and yields substantial gains on task-level tool-use performance.

\subsection{Evidence for Case-Conditioned Adaptation}\label{sec:evidence}
\begin{figure}[ht!]
    \centering
    \begin{minipage}[t]{0.49\columnwidth}
        \centering
        \includegraphics[width=\linewidth]{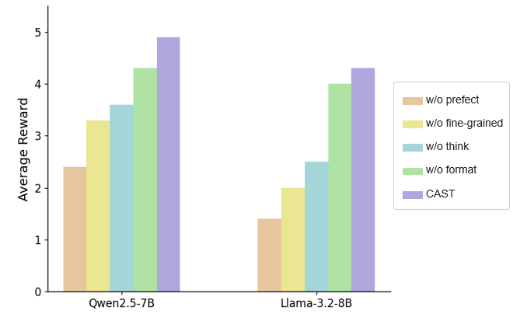}
        \refstepcounter{figure}\label{FIGN}
        \small \textbf{\figurename~\thefigure.} Ablation study of the major adaptation components.
    \end{minipage}
    \hfill
    \begin{minipage}[t]{0.49\columnwidth}
        \centering
        \includegraphics[width=\linewidth]{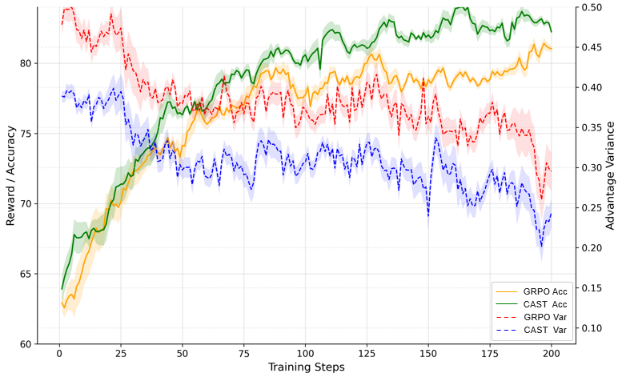}
        \refstepcounter{figure}\label{fig:training-stability}
        \small \textbf{\figurename~\thefigure.} Training accuracy and normalized advantage variance.
    \end{minipage}
\end{figure}

We evaluate CAST through ablations, budget sensitivity, and training stability. Figure~\ref{FIGN} shows that removing any major component consistently lowers average reward across backbones, indicating that CAST depends on the interaction between reasoning-side and tool-side adaptation. Table~\ref{tab:ablation} confirms this on BFCLv2. The dynamic-only variant, which keeps the adaptive reasoning-budget component without the full schema-level reward, reduces average length from 236.9 to 164.7 tokens while reaching 85.50 Non-Live AST. This confirms that the adaptive reasoning component alone can suppress unnecessary deliberation, although schema-faithful optimization is still needed for the best execution accuracy. Combining both gives the best result 88.24\% Non-Live AST at 175.4 tokens. Table~\ref{tab:complexity-threshold-revised} shows the importance of budget control. Removing the cap increases average length to 486.2 tokens and lowers BFCLv2 Overall to 86.3\%. A strict 50th-percentile cap shortens outputs to 140.8 tokens but drops ToolBench Pass to 76.5\%. The 80th-percentile setting achieves the best overall trade-off across BFCLv2 and ToolBench, supporting case-conditioned reasoning budgets for heterogeneous tool-use tasks.

\begin{table}[t]
\centering
\footnotesize
\caption{Ablation of reasoning-side and tool-side adaptation on BFCLv2.}
\label{tab:ablation}
\begin{tabular}{lcc}
\toprule
Variant & Non-Live AST $\uparrow$ & Avg.\ Reasoning Length (tok) $\downarrow$ \\
\midrule
base-GRPO    & 87.05 & 236.9 \\
dynamic-only & 85.50 & 164.7 \\
schema-only  & 86.59 & 214.5 \\
CAST         & 88.24 & 175.4 \\
\bottomrule
\end{tabular}
\end{table}

\begin{table}[t]
\centering
\footnotesize
\caption{Impact of varying reasoning length budgets on task accuracy and schema alignment.}
\label{tab:complexity-threshold-revised}
\setlength{\tabcolsep}{4pt}
\renewcommand{\arraystretch}{0.9}
\begin{tabular*}{\textwidth}{@{\extracolsep{\fill}} l S[table-format=2.2] S[table-format=2.2] S[table-format=2.2] S[table-format=3.1] @{}}
\toprule
Threshold & {\makecell{BFCLv2\\(Overall)}} & {\makecell{ToolBench\\(Pass)}} & {\makecell{ToolBench\\(Win)}} & {Length} \\
\midrule
0 (no cap)        & 86.3 & 69.9 & 65.7 & 486.2 \\
100th (loose)     & 87.5 & 80.3 & 79.6 & 240.9 \\
80th (default)    & \bfseries 88.43 & \bfseries 80.67 & \bfseries 79.43 & \bfseries 175.4 \\
50th (strict)     & 86.8 & 76.5 & 73.1 & 140.8 \\
\bottomrule
\end{tabular*}
\end{table}

Figure~\ref{fig:training-stability} shows that CAST also stabilizes RL training. Its normalized advantage variance decreases from 0.48 to 0.10, while GRPO remains around 0.21, indicating cleaner credit assignment and fewer oscillations between overthinking and underthinking.

\subsection{Case Complexity and Failure Profiles}\label{sec:analysis}
To verify that the case-derived profiles operate as intended, we analyze their impact across three dimensions: instance-level budget allocation, structural error suppression, and global curriculum organization.

\begin{figure}[t]
    \centering
    \begin{minipage}[t]{0.49\columnwidth}
        \centering
        \includegraphics[width=\linewidth]{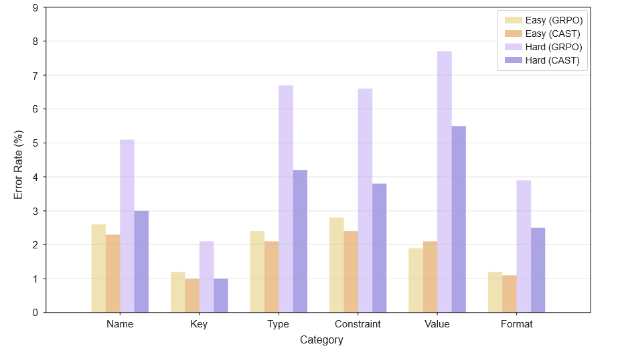}
        \refstepcounter{figure}\label{fig:agent-component}
        \small \textbf{\figurename~\thefigure.} Performance across fine-grained difficulty levels.
    \end{minipage}
    \hfill
    \begin{minipage}[t]{0.49\columnwidth}
        \centering
        \includegraphics[width=\linewidth]{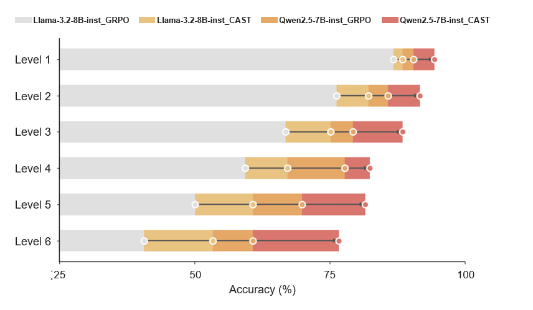}
        \refstepcounter{figure}\label{fig:error-dist}
        \small \textbf{\figurename~\thefigure.} Error-rate distribution by category.
    \end{minipage}
\end{figure}

\textbf{Adaptive Budget Allocation via Complexity Profiles.} Figure~\ref{fig:agent-component} reports execution accuracy across fine-grained difficulty levels. If the complexity profile merely acted as a length penalty, its effect would be roughly uniform across buckets or even harmful on harder cases. Instead, CAST yields only small gains on easy instances, where budget control mainly reduces verbosity, but substantially larger gains as difficulty increases. This pattern suggests that the complexity profile enables adaptive reasoning allocation: CAST shortens reasoning when additional deliberation is unnecessary, but preserves or expands the reasoning budget when compositional reasoning is required.

\textbf{Structural Error Suppression via Failure Profiles.} We next study whether schema-faithful tool optimization targets specific failure modes. Figure~\ref{fig:error-dist} separates execution errors into structural violations and localized content errors. As shown in the Reward Calculation module of Figure~\ref{fig2}, $\mathcal{R}_{\mathrm{Tool}}$ explicitly penalizes structural deviations. Accordingly, CAST reduces structural failures much more than the GRPO baseline, leaving most residual errors in localized value prediction. This result indicates that CAST improves the conversion of free-form reasoning into schema-compliant tool execution.

\begin{table}[H]
  \centering
  \caption{Effect of curriculum strategy on ToolBench.}
  \label{tab:dynamic}
  \footnotesize
  \setlength{\tabcolsep}{4pt}
  \renewcommand{\arraystretch}{0.9}
  \begin{tabular*}{\textwidth}{@{\extracolsep{\fill}} l
      S[table-format=2.1]
      S[table-format=2.1]
      S[table-format=4.1] @{}}
    \toprule
    Method &
      \multicolumn{1}{c}{Pass (\%)} &
      \multicolumn{1}{c}{Win (\%)} &
      \multicolumn{1}{c}{Length} \\
    \midrule
    No Selection & 73.2 & 69.5 & 2417.3 \\
    Two Stage    & 76.8 & 74.2 & 297.3 \\
    Hard to Easy & 68.5 & 64.3 & 426.3 \\
    Easy to Hard & \bfseries 80.7 & \bfseries 79.4 & \bfseries 175.4 \\
    \bottomrule
  \end{tabular*}
\end{table}

At the training level, Table~\ref{tab:dynamic} tests whether the case-derived complexity score $H(q)$ provides a useful curriculum signal. The easy-to-hard schedule achieves the best accuracy and the shortest outputs, with an average generation length of 175.4 tokens. In contrast, the hard-to-easy schedule performs worst and produces much longer outputs (426.3 tokens on average), suggesting that early exposure to difficult cases encourages unstable trial-and-error behavior and persistent overthinking. Overall, case-derived complexity serves as an effective signal for both local budget control and global curriculum design.

\begin{figure}[H]
    \centering
    \includegraphics[width=1\linewidth]{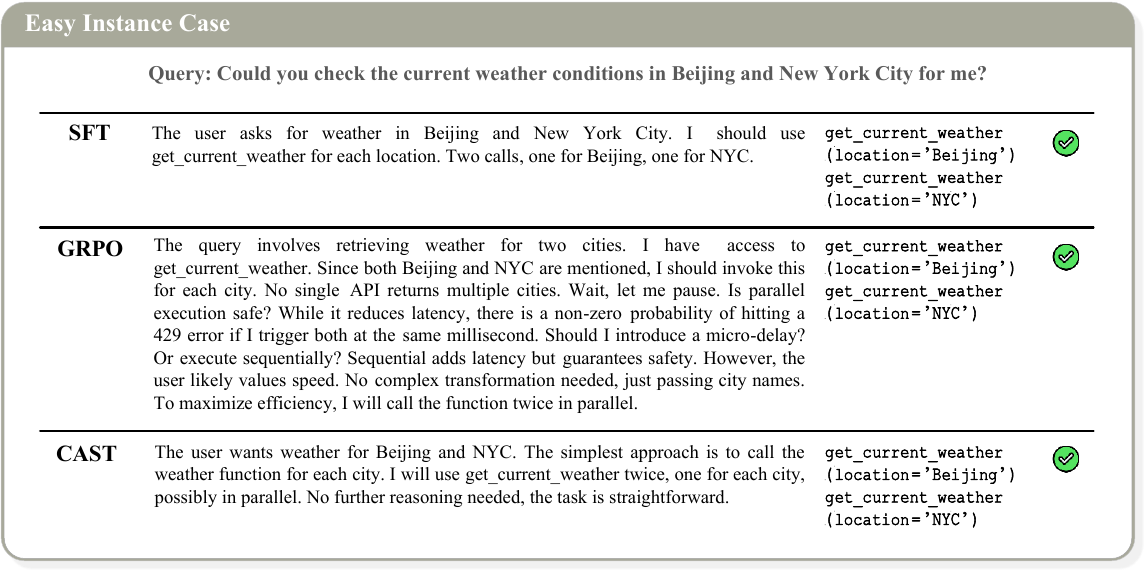}
    \caption{Easy case.}
    \label{fig:easy-instance}
\end{figure}

\begin{figure}[H]
    \centering
    \includegraphics[width=1\linewidth]{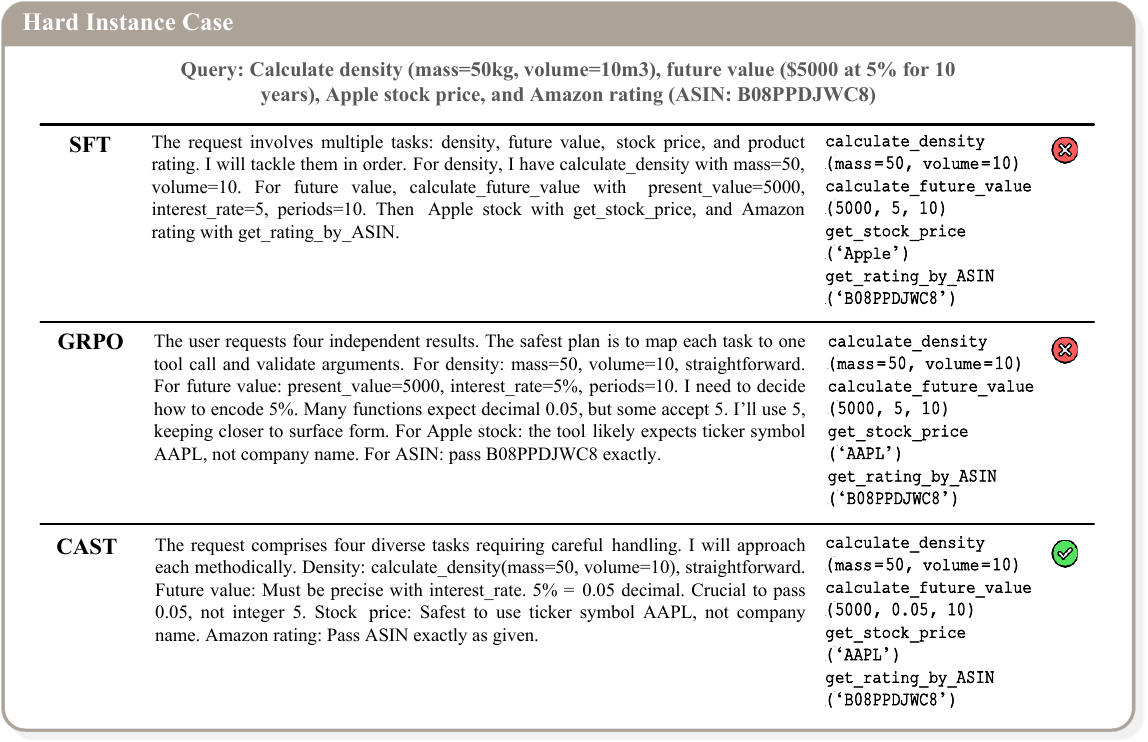}
    \caption{Hard case.}
    \label{fig:hard-instance}
\end{figure}

\subsection{Case Study}\label{sec:case_study}
Figures~\ref{fig:easy-instance} and~\ref{fig:hard-instance} illustrate CAST's adaptive reasoning-execution boundary. In the easy weather-query case, GRPO produces redundant deliberation about rate limits and execution order, whereas CAST directly emits the necessary tool calls. In the harder compositional case, SFT and GRPO copy the surface form of $5\%$ into the tool arguments, while CAST preserves enough reasoning to normalize it to $0.05$ and keeps the remaining calls structurally valid. This shows that CAST shortens reasoning when execution is straightforward and preserves it when semantic normalization is required.

\section{Conclusion}\label{sec:conclusion}
This paper revisits tool use from a case-based reasoning perspective and proposes CAST, a framework that uses signals distilled from past execution cases to guide reinforcement learning. Specifically, CAST summarizes historical trajectories through complexity and failure profiles, and uses them to regulate reasoning length and supervise schema-level tool execution. Experiments on BFCLv2 and ToolBench show that this case-conditioned calibration improves execution
accuracy, transfers to end-to-end tool-use success, and reduces unnecessary reasoning, with the largest gains on more complex cases. Long-horizon planning remains challenging, but the results suggest that case-derived supervision provides a practical basis for improving both reliable tool calling and downstream task completion. A natural next step is to extend this framework to richer case memories, stronger retrieval and reuse, and more interactive agent settings.

\clearpage
\begin{credits}
\subsubsection{\ackname} This work was partially supported by the National Natural Science Foundation of China (U2336204), Chengdu Industrial Chain Collaborative Innovation Project (Grant No. 2025-XT00-00017-GX) and the Open bidding for selecting the best candidates of Sichuan Provincial Department of Science and Technology (2024YFCY0003).
\subsubsection*{Generative AI Disclosure}
LLMs were used exclusively for stylistic refinement of the manuscript. All textual content was initially drafted in full by the authors and subsequently polished with the assistance of LLM-based tools, including ChatGPT and Gemini. All scientific contributions, technical methods, ideas, and core results presented in this work are entirely the original work of the authors.
\end{credits}

\bibliographystyle{splncs04}
\bibliography{bibliography}

\end{document}